\documentclass[12pt]{article}
\usepackage{epsfig, epsf, graphicx}
\usepackage{pstricks, pst-node, psfrag}
\usepackage{amssymb,amsmath,bm}
\usepackage{verbatim,enumerate}
\usepackage{rotating, lscape,natbib}
\usepackage{setspace}
\usepackage{soul}
\usepackage{tikz}
\usetikzlibrary{arrows,calc,tikzmark}
\usepackage{hyperref}
\usepackage{float}
\usepackage{caption}
\usepackage{subcaption}
\usepackage{animate}
\usepackage{mathtools}
\usepackage{multimedia}
\usepackage{comment}
\usepackage{enumitem}
\usepackage{url}
\usepackage[utf8]{inputenc}
\usepackage[english]{babel}
\usepackage{color}
\usepackage{verbatim}
\usepackage{algorithm}
\usepackage{algpseudocode}
\usepackage{multirow}

\def\boxit#1{\vbox{\hrule\hbox{\vrule\kern6pt
          \vbox{\kern6pt#1\kern6pt}\kern6pt\vrule}\hrule}}

\def\bse{\begin{eqnarray*}}
\def\ese{\end{eqnarray*}}
\def\be{\begin{eqnarray}}
\def\ee{\end{eqnarray}}
\def\bq{\begin{equation}}
\def\eq{\end{equation}}
\def\bse{\begin{eqnarray*}}
\def\ese{\end{eqnarray*}}

\begin{document}

\def\algbackskip{\hskip-\ALG@thistlm}
\makeatother

\thispagestyle{empty} \baselineskip=28pt \vskip 5mm
\begin{center} {\Huge{\bf Towards Black-Box Parameter Estimation}}
	
\end{center}

\baselineskip=12pt \vskip 10mm

\begin{center}\large

Amanda Lenzi\footnote[1]{
\baselineskip=10pt School of Mathematics, University of Edinburgh,
Edinburgh EH9 3FD, Scotland, United Kingdom.} and
Haavard Rue\footnote[2]{
\baselineskip=10pt Statistics Program,
King Abdullah University of Science and Technology,
Thuwal 23955-6900, Saudi Arabia.}
\end{center}

\baselineskip=17pt \vskip 10mm 

\begin{center}
{\large{\bf Abstract}}
\end{center}
Deep learning algorithms have recently been shown to be a successful tool in estimating parameters of statistical models for which simulation is easy, but likelihood computation is challenging. 
This is achieved by sampling a large number of parameter values from a distribution, which is typically chosen to be non-informative and cover as much of the parameter space as possible.
However, for high-dimensional and large parameter spaces, covering all possible reasonable parameter values is infeasible.
We propose a new sequential training procedure that reduces simulation cost and guides simulations toward the region of high parameter density based on estimates of the neural network and the observed data. 
Our following proposal aims to fit time series models to newly collected data at no cost using a pre-trained neural network with simulated time series of a fixed length. 
These approaches can successfully estimate and quantify the uncertainty of parameters from non-Gaussian models with complex spatial and temporal dependencies.
The success of our methods is a first step towards a fully flexible automatic black-box estimation framework.


\baselineskip=14pt

\par\vfill\noindent
\par\medskip\noindent
{\bf Short title}: Black-box Estimation

\clearpage\pagebreak\newpage \pagenumbering{arabic}
\baselineskip=26pt

\section{Introduction}\label{sec:intro}

Statistical modeling consists of first devising stochastic models for phenomena we want to learn about and then to relate those models to data.
These stochastic models have unknown parameters and the second step boils down to estimating these parameters from data through the likelihood function. 
However, there might be a discrepancy between these two steps, as models for describing mechanisms aim for scientific adequacy rather than computational tractability.
Indeed, as soon as we move away from Gaussian processes as the canonical model for dependent data, likelihood computation becomes effectively impossible, and inference is too complicated for traditional estimation methods.
Consider, for instance, datasets from finance or climate science, where skewness and jumps are commonly present and calculating the likelihood in closed form is often impossible, ruling out any numerical likelihood maximization and Bayesian methods. 
Yet it is computationally inexpensive to simulate from those models given parameter values, and the question becomes whether the simulations look like the data.

Much effort has been directed toward the development of approximate parameter estimation methods, often referred to as indirect inference \citep{gourieroux1993indirect}, likelihood-free inference \citep{grelaud2009abc, gutmann2016bayesian}, simulation-based inference \citep{nickl2010efficient} or synthetic likelihood \citep{wood2010statistical}; for an overview, see, for example, the review by \citet{hartig2011statistical, cranmer2020frontier}.
The typical assumption by the different methods is that exact likelihood evaluation is hard to obtain but it is easy to simulate from the model given the parameter values, and the basic idea is to identify the model parameters which yield simulated data that resemble the observed data.
The most common in this umbrella is arguably approximate Bayesian computation (ABC) \citep{fearnhead2012constructing, frazier2018asymptotic, sisson2018handbook}, which avoids evaluating intractable likelihoods by matching summary statistics from observations with those computed from simulated data based on parameters drawn from a predefined prior distribution. 
The likelihood is approximated by the probability that the condition $\gamma(x_{\mbox{sim}}, x_{\mbox{obs}}) < \epsilon$ is satisfied, where $\gamma$ is some distance measure and the value of $\epsilon$ is a trade-off between sample efficiency and inference quality.
In simpler cases, sufficient statistics are used as they provide all the information in the data, however, for complex models they are unlike to exist and it is not obvious which statistics will be most informative.
Several works have proposed procedures for designing summary statistics \citep{fearnhead2012constructing, jiang2017learning}, and a comparison of likelihood-free methods with and without summary statistics have been empirically tested in \citet{drovandi2022comparison}.
Despite its popularity, ABC is not scalable to large numbers of observations since inference for new data requires repeating most steps of the procedure. 

A recent line of research on likelihood-free inference uses deep learning to estimate parameters of statistical models.
The first work to propose neural networks-based estimators in the statistical community focused on parameters of spatial covariance functions in Gaussian processes \citep{gerber2021fast}. 
They showed that convolutional neural networks (CNNs) can learn the mapping between data and parameters and had similar estimation accuracy and a considerable reduction in computational time compared to classical maximum likelihood estimators.
Based on the same idea, \cite{lenzi2021neural} estimated parameters of models for spatial extremes for which the likelihoods are intractable and, therefore, MLEs are unavailable. 
A modified parametric bootstrap approach was introduced to quantify the uncertainty in these estimators. 
Variants to those methods to incorporate replicated data in the estimation \citep{sainsbury2022fast}, irregular spatial data \citep{sainsbury2023neural}, and censoring information \citep{jordan2023neural} were successfully recently introduced.
Estimation in high dimensions and large parameter spaces is still an open question since constructing training data that covers all possible reasonable parameter values becomes quickly infeasible. 
An inevitable drawback of the current approaches is the estimator's bias towards the parameter region of the training data. 
Whereas previous work simulated training data either with parameter values around the truth or based on inexact likelihood estimates, these methods are expensive and unrealistic.
In this work, we solve the bias issue of previous methods with an automatic iterative approach that modifies the training data using arbitrary, dynamically updated distribution parameters until it reaches the parameter region corresponding to the actual data.
The proposed inference mechanism automatically performs parameter estimation without restrictive assumptions about the generating process, knowledge from experts, or computationally expensive preliminary steps. 

Various approaches have been proposed for guiding simulations by making use of the observed data. 
The sequential training procedure proposed here falls into a broader class of methods that seek to update the prior distribution to alleviate the curse of dimensionality that comes with having to explore a large volume of data when the number of parameters is large. 
Within ABC implementations, a Sequential Monte Carlo procedure was used to guide simulations based on previously accepted parameters \citep{sisson2007sequential, beaumont2009adaptive, bonassi2015sequential}.
\citet{jarvenpaa2019efficient, lueckmann2019likelihood} updated simulations that reduce the Bayesian uncertainty in the posterior estimate. 
A sequential estimation procedure selected future simulations by proposing parameters from preliminary approximations to the posterior \citep{lueckmann2017flexible, papamakarios2016fast}.
Sequential approaches have been considered in the context of neural networks for learning a model of the likelihood in the region of high posterior density \citet{papamakarios2019sequential}, but only for toy examples and small datasets. 
Our approach stands out from previous methods as it is aimed at point parameter estimation in intractable statistical models with large and complex datasets.
We leverage our statistical knowledge about scaling data and parameters to reduce computational costs and improve convergence when training the neural networks.
We first illustrate the new approach on a Gaussian toy example, where the intended coverage probability of the estimator is known, and further for modeling spatial extremes. 
Similarly to \citet{papamakarios2019sequential}, which trains autoregressive flows on all simulations obtained up to each round rather than training only with simulations from the latest round, we broaden the range of the training by reusing training data from previous iterations.
Consequently, most of the training data in later rounds still come from the most probable parameter regions, and the added simulations from previous rounds broaden the range of searches at no computational cost.
We leverage the ideas of improving the estimator by updating the prior in multiple rounds. 
Since our approach is designed to yield precise point estimates instead of approximating posterior distributions, the loss function in the neural network remains independent of the current proposed prior.
We apply the modified parametric bootstrap technique in \citet{lenzi2021neural} such that in each round, the new approach is able to quantify both the bias and variance of the estimators.
Our experimental results show that the sequential approach dramatically reduces the bias of an initial guess and eventually approximates the actual parameter quite accurately, even when the initial training does not contain the truth.

Our next contribution is designed to handle estimation for different data sizes originating from the same model. 
This is achieved by training a neural network only once on an extensive database and replicating the observations as needed to achieve the training data length before processing them on the pre-trained neural network. 
We adjust for the underestimation in the uncertainty of the replicated series by rescaling its sampling variance to match the original series variance.
We apply this strategy to estimate parameters of non-Gaussian stochastic volatility models, which are widely used in finance. 
In such applications, where early access to results may carry a premium, our deep neural network (DNN) estimator is particularly advantageous since the network is trained beforehand, and estimates are obtained instantaneously when new data becomes available. 
This example shows that our estimator is well calibrated, with uncertainty quantification closely matching those from the state-of-the-art Integrated Nested Laplace Approximation (INLA) approach \citep{rue2009approximate}.


The remainder of this paper is organized as follows. 
First, Section~\ref{sec:parm-est} outlines the methodologies we develop for designing training data to train DNNs, along with some practical considerations. 
In Section~\ref{sec:iter}, we introduce the construction of the automatic iterative approach and conduct simulation studies for Gaussian i.i.d and spatial extremes model, whereas in Section~\ref{sec:ts}, we describe our unified database approach applied to time series data from an intractable model.
In Section~\ref{sec:conclusion}, we conclude and summarize avenues for future research.

\section{Parameter estimation with DNNs} \label{sec:parm-est}

\subsection{Background} \label{sec:parm-methods}

Consider a dataset  $\mathbf{x}_0 \in \mathbb{R}^J$ of observations generated from $\mathbb{P} \in \mathcal{P}(\mathcal{X})$, where $\mathbb{P}$ is a Lesbegue measure and $\mathcal{P}(\mathcal{X})$ denote the set of all Borel probability measures on the sample space $\mathcal{X}$.
To describe such a process, it is common practice to assume a statistical model $\mathbb{P}_{\boldsymbol{\theta}_0} \in \mathcal{P}(\mathcal{X})$ with probability density function $p(\cdot; \boldsymbol{\theta}_0)$ parameterized by a finite number of parameters $\boldsymbol{\theta}_0 \subset \Theta \in \mathbb{R}^P$, which is estimated using the observations through the log-likelihood function $l(\boldsymbol{\theta}_0; \mathbf{x}_0) \equiv \mbox{log}\{p(\mathbf{x}_0; \boldsymbol{\theta}_0)\}$.

Highly structured data coming from a high-dimensional $\mathcal{X}$ 
are often related to intractable or computationally demanding likelihoods, but simulating data from $p$ for given parameters is usually trivial.
Recently, parameter estimation using DNNs have opened doors to solving previously intractable statistical estimation problems.
The key to efficiency is to avoid altogether learning likelihood functions and directly learn the mapping between data and parameters through DNNs by carrying out simulations.
To formulate the problem, let $\mathbf{x}_n \in \mathbb{R}^J$ be a simulated sample from $p(\cdot; \boldsymbol{\theta}_n)$ with given parameters $\boldsymbol{\theta}_n  \in \mathbb{R}^P$. 
Then, the mapping from $\mathbf{x} = (\mathbf{x}_1, \ldots, \mathbf{x}_N)^\top$ onto $\boldsymbol{\theta} = (\boldsymbol{\theta}_1, \ldots, \boldsymbol{\theta}_N)^\top$ is learned by adjusting the weights $\textbf{w}$ and biases $\textbf{b}$, denoted by $\phi = (\textbf{w}, \textbf{b})^{\top}$ of a DNN $\mathcal{F}_{\phi}$, such that
\begin{equation}
    \mathcal{F}_{\phi} : \mathbf{x} \mapsto \boldsymbol{\theta} ; \quad
        \hat{\boldsymbol{\theta}} = \mbox{argmin}_{\phi} \mathcal{L}\{\boldsymbol{\theta}, \mathcal{F}_{\phi}(\mathbf{x})\}.
    \label{eq:param-nn}
\end{equation}
Optimizing \eqref{eq:param-nn} with respect to $\phi$ requires the minimization of the loss function $\mathcal{L}$, which is chosen to reduce the error in prediction for a given output $\boldsymbol{\theta}$ and simulated data $\mathbf{x}$.
A popular choice in regression problems is the mean squared error (MSE)
\begin{equation}
    \mathcal{L}(\phi; \boldsymbol{\theta}, \mathbf{x}) = \mathbb{E} \{\boldsymbol{\theta} - \mathcal{F}_{\phi}(\mathbf{x})\}^2.
    \nonumber
    \label{eq:loss}
\end{equation}
Often, no closed-form solutions can be derived for the optimization in \eqref{eq:param-nn}, and advanced numerical optimizers built around batch gradient descent methods are employed \citep{kingma2014adam}.
Finally, once the DNN has been trained, one can use the estimated $\hat{\phi}$ to plug in $\mathbf{x}_0$ into the trained DNN and retrieve $\mathcal{F}_{\hat{\phi}}(\mathbf{x}_0)$, which will then output parameter estimates of interest $\hat{\boldsymbol{\theta}}_0$.

\subsection{Transformations to data and parameters} \label{sec:transf}

Here, we detail our rationale for choosing transformations to data and parameters and make the problem more palatable for the DNN. 
The key is to use our statistical knowledge of intrinsic data and parameter properties to leverage estimation.
For instance, the quadratic loss in \eqref{eq:loss} is optimal for outputs with constant mean and variance that are a real-valued function of the inputs with Gaussian distributed noises. 
If these assumptions are met, the estimator $\mathcal{F}$ retains the desired properties, such as the minimum variance and fast convergence, as the gradient reduces gradually for relatively small errors \citep{friedman2001elements}.
Therefore, reparametrization related to scaling should aim for a constant variance for different parameter values. 

When minimizing the loss in \eqref{eq:param-nn}, $\phi$ is usually initialized to random values and updated via an optimization algorithm, such as stochastic gradient descent based on training data. 
As is usually the case, the geometry of the surface that has to be optimized will be complicated and smooth due to an ample search space and noisy data, and using the raw data will likely result in slow and unstable convergence. 
To remedy this problem and improve the algorithm's stability, one should aim for properties such as symmetric and unbounded distributions, orthogonal parameters, and constant Fisher information. 
The logarithm and square root transformations often used in time series problems are examples of desirable change that affects the distribution shape by reducing skewness, stabilizing the variance, and simultaneously avoiding boundaries. 
Parameterization with meaningful interpretations such as mean and variance should be preferred over directly using distribution parameters.

In Sections~\ref{sec:iter} and \ref{sec:ts}, we will use transformation within a DNN pipeline to estimate parameters of models for Gaussian data, spatial extremes, and non-Gaussian stochastic volatility models. We show that whereas these precautions are helpful in simple Gaussian examples, they are indispensable in complex models such as for spatial extremes. 

\subsection{Designing training data}

Recall that the first step for optimizing \eqref{eq:param-nn} is generating pairs of training data $(\boldsymbol{\theta}_n, \mathbf{x}_n)^N_{n=1}$.
The main challenge here is to generate training data that correspond to configurations covering the parameter domain of the observations, which is unknown. 
Since $\boldsymbol{\Theta}$ is often unbounded and it is impossible to simulate over the entire domain. Not introducing an appropriate structure or prior scientific knowledge will lead to inaccurate training data and, thus, erroneous estimation.

In the context of intractable likelihoods for which MLEs are unavailable, \citet{lenzi2021neural} proposed to simulate training data based on informative parameter estimates from approximate maximum likelihood methods fit to spatial extremes data.
However, obtaining these estimates for every new dataset becomes problematic if likelihood estimation is slow or not feasible in the first place.
Here, we propose two different strategies to deal with this challenge. The intuition behind these method goes as follow:
\begin{enumerate}[label=(\Alph*)]
    \item {\textit{A fully automatic iterative approach:}}
        Promising regions in  $\boldsymbol{\Theta}$ are found sequentially. The DNN is initially trained with $\boldsymbol{\theta}$ based on a crude guess (e.g., from a simpler model). The trained DNN then receives $\mathbf{x}_0$ and outputs $\hat{\boldsymbol{\theta}}_0$, which is used to simulate bootstrapping samples $\mathbf{x}_b \sim  p(;\hat{\boldsymbol{\theta}}_0)$. Next, $\mathbf{x}_b$ is fed into the trained DNN to output bootstrapping samples $\hat{\boldsymbol{\theta}}_b$. The spread of $\hat{\boldsymbol{\theta}}_b$ and its distance from $\hat{\boldsymbol{\theta}}_0$ are used to guide simulations for training the DNN. 
        Empirical results show a higher concentration of training data and bootstrapping samples around the true parameters after a few iterations, hence a better approximation of $\boldsymbol{\theta}$.
    
    \item \textit{A general unified database approach for time series:} 
    Computation efficiency is achieved by training the DNN in advance and reusing it to estimate newly collected data for free. 
    The pre-training is based on an extensive database comprising simulated time series data $\mathbf{x}$ and corresponding parameters $\boldsymbol{\theta}$. 
    Next, parameter estimates from new data of different lengths are obtained by replicating the observations to achieve the size of $\mathbf{x}$. 
\end{enumerate}

The details on the building blocks of the frameworks in (A) and (B), along with the necessary adjustments for different applications, are given in Sections~\ref{sec:iter} and~\ref{sec:ts}.

\section{A fully automatic iterative approach}
\label{sec:iter}

\subsection{General framework}

We now use the notation in Section~\ref{sec:parm-est} to describe an algorithm that sequentially samples training data until it reaches the correct parameter regions from the data.
Our algorithm is initialized with simulated pairs ($\boldsymbol{\theta}_n, \mathbf{x}_n)^N_{n=1}$, where the elements in $\boldsymbol{\theta}_n \in \mathbb{R}^P$ are draw from $P$ independent Uniform distributions, each bounded below by $(a_{1,p})^P_{p=1}$ and above by $(a_{2,p})^P_{p=1}$, while $\mathbf{x}_n \in \mathbb{R}^J$ is data simulated with $\boldsymbol{\theta}_n, n=1, \ldots, N$. 
Intervals $(a_{1,p}, a_{2,p})^P_{p=1}$ may or may not contain the true parameter $(\theta_{0,p})^P_{p=1}$.
Whereas good initial guesses of $(a_{1,p})^P_{p=1}$ and $(a_{2,p})^P_{p=1}$ are not essential here, most models allow for some data-driven estimates, e.g., based on simplified Gaussian assumptions. 
Next, a DNN is trained with ($\boldsymbol{\theta}_n, \mathbf{x}_n)^N_{n=1}$, and then used to retrieve estimates $\hat{\boldsymbol{\theta}}_0 \in \mathbb{R}^P$ when fed with observations $\mathbf{x}_0 \in \mathbb{R}^J$. 

The main idea is to then dynamically update the training data ($\boldsymbol{\theta}_n, \mathbf{x}_n)^N_{n=1}$ by changing the values of $(a_{1,p})^P_{p=1}$ and $(a_{2,p})^P_{p=1}$ based on information on whether $\hat{\theta}_{0,p}$ is underestimating or overestimating ${\theta}_{0,p}$.
For instance, if $\hat{\theta}_{0,p}$ is close to the upper boundary of the training data for a specific $p$, then new training samples should be expanded to contain data outside of that boundary.
Information on the accuracy of $(\hat{\theta}_{0,p})^P_{p=1}$ is obtained by sampling new data $\mathbf{x}_b = (\mathbf{x}_b^1, \ldots, \mathbf{x}_b^B)^{\top}$ with $\mathbf{x}_b \in \mathbb{R}^{J \times B}$ using $\hat{\boldsymbol{\theta}}_0$, and feeding these data into the initially trained DNN producing a bootstrapped sample $\hat{\boldsymbol{\theta}}_b = (\hat{\boldsymbol{\theta}}_b^1, \ldots, \hat{\boldsymbol{\theta}}_b^B)^{\top}, \hat{\boldsymbol{\theta}}_b \in \mathbb{R}^{P \times B}$.
For each $p$, we then update $a_{1,p}$ and $a_{2,p}$ to values in a neighborhood of $\hat{\theta}_{0,p}$, where the neighborhood region is defined by the size of the bias between  $\hat{\theta}_{0,p}$ and $\tilde{\theta}_{p}$, where $\tilde{\theta}_{p}$ is the median of $\theta^1_{p}, \ldots, \theta^B_p$, and the neighborhood width depends on the quantiles of the bias between the fitted value and the bootstrapped sample: $\mathcal{Q}^{\alpha}_p(\hat{\theta}_{0,p} - \theta^1_{p}, \ldots, \hat{\theta}_{0,p} -\theta^B_p)$, where $\alpha$ is a quantile.
The algorithm stops when the bias between $\hat{\theta}_{0,p}$ and $\tilde{\theta}_{p}$ is sufficiently small compared to the standard deviation of $(\theta^1_{p}, \ldots, \theta^B_p)^{\top}$, which we denote by $S_{p}$, for all $p$.
In more detail, the algorithm is as follows.
    \begin{algorithm}[htb!]
    \caption{Iterative procedure}
    \textbf{Need:} Observations $\mathbf{x}_0 \in \mathbb{R}^{J}$ from a distribution $p, \boldsymbol{\theta}_0 \in \mathbb{R}^{P}$ and a neural network $\mathcal{F}_{\phi}(\cdot)$ \\
    Pick $\gamma \in (0,1)$, $a_{1,p}$ and $a_{2,p}, p=1, \ldots, P,$
    \begin{algorithmic}[1]
    \While{$\mbox{bias}(\hat{\theta}_{0,p}, \tilde{\theta}_{b,p}) > \gamma \times S_{p}$, for all $p$} 
    \State Sample $\theta_{n,p} \sim \mbox{Unif}(a_{1,p}, a_{2,p}), n = 1, \ldots, N$
    \State Simulate $\mathbf{x}_n \sim p(;\boldsymbol{\theta}_n), n = 1, \ldots, N$
    \State Train $\mathcal{F}_{\phi}(\mathbf{x})$ and obtain $\hat{\boldsymbol{\theta}}_0$ from $\mathcal{F}_{\hat{\phi}}(\mathbf{x}_0)$
    \State Simulate $\mathbf{x}_b \sim p(;\hat{\boldsymbol{\theta}}_0)$ and obtain $\hat{\boldsymbol{\theta}}_{b}$ from $\mathcal{F}_{\hat{\phi}}(\mathbf{x}_b), b = 1, \ldots, B$
    \State $a_{1,p} = \hat{\theta}_{0,p} +  \mbox{bias}(\hat{\theta}_{0,p}, \tilde{\theta}_{p}) -  \mathcal{Q}^{0.05}_p(\hat{\theta}_{0,p} - \theta^1_{p}, \ldots, \hat{\theta}_{0,p} -\theta^B_p)$
    \State $a_{2,p} = \hat{\theta}_{0,p} +  \mbox{bias}(\hat{\theta}_{0,p}, \tilde{\theta}_{p}) +  \mathcal{Q}^{0.975}_p(\hat{\theta}_{0,p} - \theta^1_{p}, \ldots, \hat{\theta}_{0,p} -\theta^B_p)$
    \EndWhile
   \end{algorithmic}
   \label{alg1}
    \end{algorithm}
    

Small values of $\gamma$ in line 1 of Algorithm~\ref{alg1} will make the algorithm run longer, since it requires  $\hat{\theta}_{0,p}$ to be closer to $\tilde{\theta}_{b,p}$ relative to the spread of $\theta^1_{p}, \ldots, \theta^B_p$. 
At each iteration, line 5 automatically provides uncertainty quantification of $\hat{\boldsymbol{\theta}}_0$ through $\hat{\boldsymbol{\theta}}_{b}$.
This step works as a modified and more efficient parametric bootstrap method since it uses the previously trained DNN, and no model fitting is required to produce $\hat{\boldsymbol{\theta}}_{b}$.
One can use these samples to compute quantities of interest, such as confidence intervals and coverage, and check the overall appropriateness of the method.
Here, we use them to quantify the accuracy of the current iteration and to update the training data for the next round (see lines 6 and 7).


Parameter values $\boldsymbol{\theta}$ are usually in the transformed scale, and we continuously sample training data such that the values in this transformed scale are uniformly distributed (see line 2 of Algorithm~\ref{alg1}) rather than applying transformations after the training data have been generated to train the DNN.
The former would produce regions of scarcity in $\Theta$, and results for testing data within the underrepresented values would not be optimal.
Indeed, the optimization inside the DNN will perform best if the training data have no significant gaps between values, a problem also called imbalanced data in classification problems \citep{murphey2004neural}.

In Section~\ref{sec:iid}, we estimate the parameters of an i.i.d Gaussian model as a proof-of-concept, whereas, in Section~\ref{sec:maxstab}, we consider a spatial-extremes setting and estimate the parameters of the Brown-Resnick max-stable process with an intractable likelihood.
This procedure supports a wide range of likelihoods with fixed and random effects, and distributions other than the Uniform could also have been used in line 2 of Algorithm~\ref{alg1}.

\subsection{I.i.d. data}
\label{sec:iid}

Although classical inference for the models considered in this section is straightforward, they allow us to compare our estimates' accuracy and uncertainty with MLEs. 
For applications where our method is of practical interest, see Section~\ref{sec:maxstab}.
In what follows, we look at three problems of increasing complexity from parameters of Gaussian distributions: the logarithm variance (single parameter), the mean and the logarithm variance (two orthogonal parameters), and the first moment and logarithm of the second moment (two highly dependent parameters).
With these examples, we aim to empirically illustrate that our framework:
(1) approaches the MLE even when the initial training data is relatively far from the actual value, and
(2) reaches the truth quicker when using meaningful parametrizations and orthogonal parameters.

Consider i.i.d. observations $\mathbf{x}_0 \in \mathbb{R}^J$ from a Gaussian distribution $\mathcal{N}(\mu,\sigma_0^2)$. 
We find that $\gamma=0.3$ in line 1 of Algorithm~\ref{alg1} is enough to provide good estimation accuracy without overly increasing computational cost. 
Algorithm~\ref{alg2} shows the steps of our procedure for the i.i.d. case (see Algorithm~\ref{alg1} for the general case), whereas some practical aspects are discussed in what follows. 

    \begin{algorithm}[htb!]
    \caption{Iterative procedure for i.i.d. data}
    \textbf{Need:} Observations $\mathbf{x}_0 \in \mathbb{R}^{J}$ from a distribution $p, \boldsymbol{\theta}_0 \in \mathbb{R}^{P}$ and a neural network $\mathcal{F}_{\phi}(\cdot)$ \\
    Pick $a_{1,p}$ and $a_{2,p}, p=1, 2$
    \begin{algorithmic}[1]
    \While{$\mbox{bias}(\hat{\theta}_{0,p}, \tilde{\theta}_{b,p}) > 0.3 \times S_{p}$, for all $p$}  
    \State Sample $\theta_{n,p} \sim \mbox{Unif}(a_{1,p}, a_{2,p}), n = 1, \ldots, N$
    \State Simulate $\mathbf{x}_n^* \sim p(;\boldsymbol{\theta}_n), n = 1, \ldots, N$
        \State Train $\mathcal{F}_{\phi}(\mathbf{x})$ and obtain $\hat{\boldsymbol{\theta}}_0$ from $\mathcal{F}_{\hat{\phi}}(\mathbf{x}_0)$
    \State Simulate $\mathbf{x}_b \sim p(;\hat{\boldsymbol{\theta}}_0)$ and obtain $\hat{\boldsymbol{\theta}}_{b}$ from $\mathcal{F}_{\hat{\phi}}(\mathbf{x}_b), b = 1, \ldots, B$
    \State $a_{1,p} = \hat{\theta}_{0,p} +  \mbox{bias}(\hat{\theta}_{0,p}, \tilde{\theta}_{p}) -  \mathcal{Q}^{0.05}_p(\hat{\theta}_{0,p} - \theta^1_{p}, \ldots, \hat{\theta}_{0,p} -\theta^B_p)$
    \State $a_{2,p} = \hat{\theta}_{0,p} +  \mbox{bias}(\hat{\theta}_{0,p}, \tilde{\theta}_{p}) +  \mathcal{Q}^{0.975}_p(\hat{\theta}_{0,p} - \theta^1_{p}, \ldots, \hat{\theta}_{0,p} -\theta^B_p)$
    \State Increase $N$ by $5\%$
    \EndWhile
   \end{algorithmic}
   \label{alg2}
    \end{algorithm}

\paragraph{Multi-layer perceptron (MLP)} 
A perceptron is a single neuron model, and MLPs are the classical type of neural network comprised of one or more layers of several neurons. It takes 1D vectors as the input and learns nonlinear relationships between inputs and outputs, making it a suitable choice for our i.i.d. regression problem.
Due to the simplicity of this toy example, we find that a small MLP with a single hidden layer and 50 hidden units is enough to near the mapping between data and parameters.
We consider an MLP for $\mathcal{F}_{\phi}$ taking output in $\mathbb{R}^P$ and input in $\mathbb{R}^J$.

\paragraph{Progressively increasing training accuracy} 
Our algorithm uses fewer training samples when estimation uncertainty is larger, at the beginning of the algorithm, and more samples towards the end, when more precision is required.
When the estimates are close to stabilizing, and the uncertainty has decreased, the number of samples in the training data is set to increase by $5\%$ (see line 8 of Algorithm~\ref{alg2}). 

\begin{figure}[htb!]
	\centering
      	\includegraphics[width=0.38\textwidth]{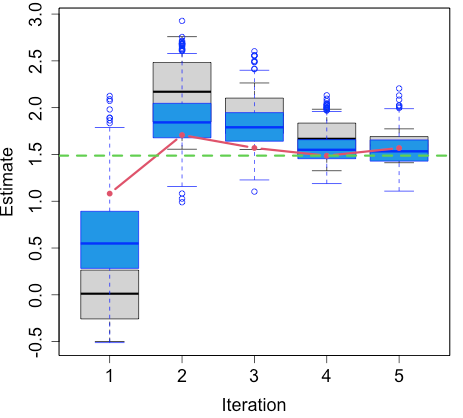}
      	\caption{Boxplots of training data (grey boxes), fitted values (red line), and bootstrapped samples (blue boxes) as iterations progress for estimating $\mbox{log}(\sigma_0^2) = 1$ from a zero-mean Gaussian distributed sample of size $J=20$ using Algorithm~\ref{alg2}.
    The horizontal dashed line corresponds to the MLE. 
 Training datasets were initially simulated in the uniform interval, with $N=10000$ samples. }
		\label{fig:gaussvar}
\end{figure}

\paragraph{Results for a single parameter} 
Figure~\ref{fig:gaussvar} displays the results for estimating $\theta_0 \equiv \mbox{log}(\sigma_0^2) = 1$ when the mean is known using Algorithm~\ref{alg2}.
We set $N=10000$ and uniformly generate training output samples $\{\theta_n\}_{n=1}^{N} \sim {\mbox{Unif}}(-2, 1)$ and corresponding inputs $\{\mathbf{x}_{n}\}_{n=1}^{N} \sim N\{1, \mbox{exp}(\theta_n)\}$.
The grey boxes in this figure are the training data at each iteration, whereas fitted values are represented by the red line, with the blue boxes showing bootstrapped estimates for $B=10000$. 
After five iterations, the algorithm approaches the MLE (green dashed line) with low uncertainty (see narrow blue boxes).
To quantify the appropriateness of our method, we compare $95\%$ central intervals from the bootstrapping estimates with the same interval from the empirical variance in the data. 
The $95\%$  interval provides adequate uncertainty of the MLP estimates, with bootstrap interval on $(0.84, 2.91)$ compared to $(0.83, 2.95)$  for the intended coverage probability of the MLE.


\begin{figure}[htb!]
	\centering
			\begin{subfigure}[b]{0.49\textwidth}
			\caption*{\scriptsize{(a) Mean and transformed variance}}
		\includegraphics[width=1\textwidth]{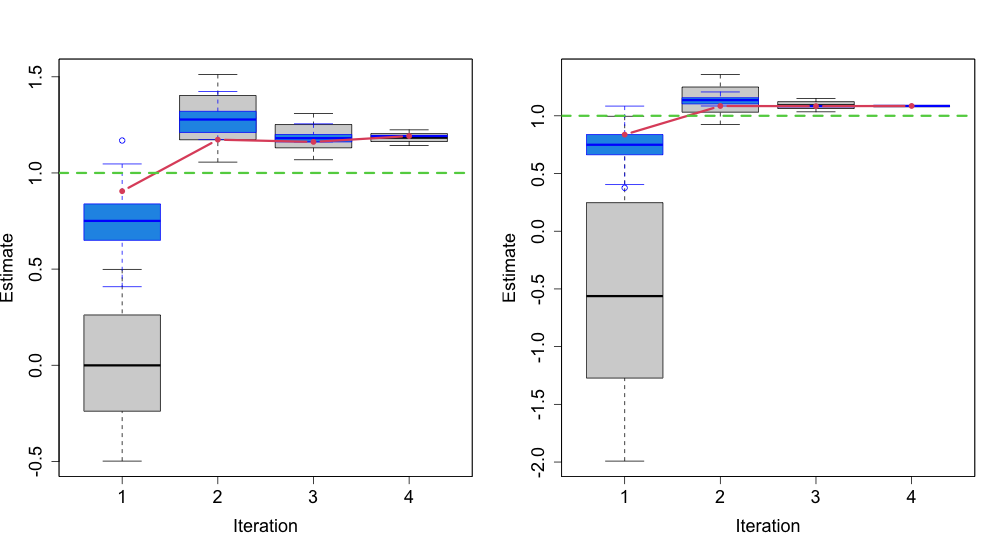}
	\end{subfigure}
				\begin{subfigure}[b]{0.49\textwidth}
							\caption*{\scriptsize{(b) Mean and variance}}
		\includegraphics[width=1\textwidth]{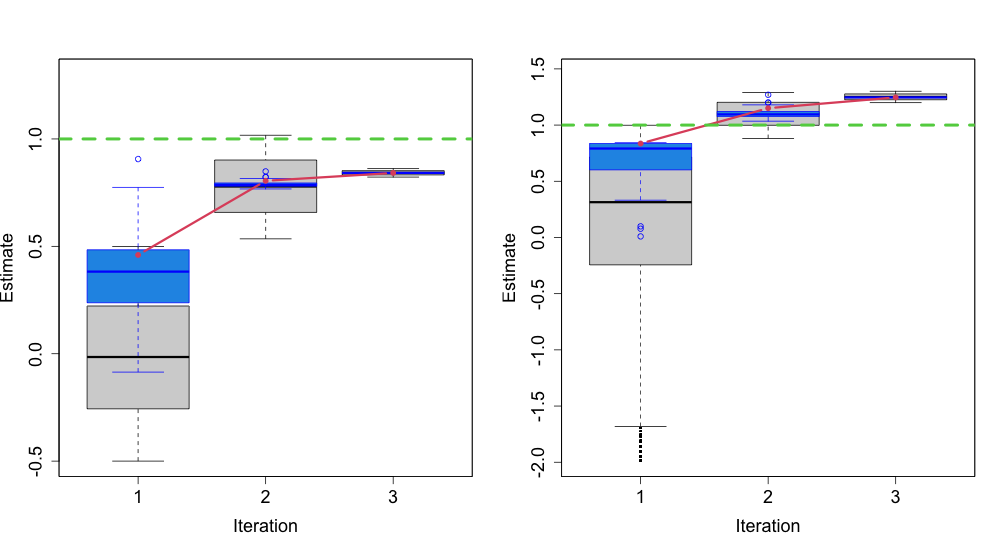}
	\end{subfigure}
	\\
     \begin{subfigure}[b]{0.49\textwidth}
      			\caption*{\scriptsize{(c) First and second moment transformed}}
		\includegraphics[width=1\textwidth]{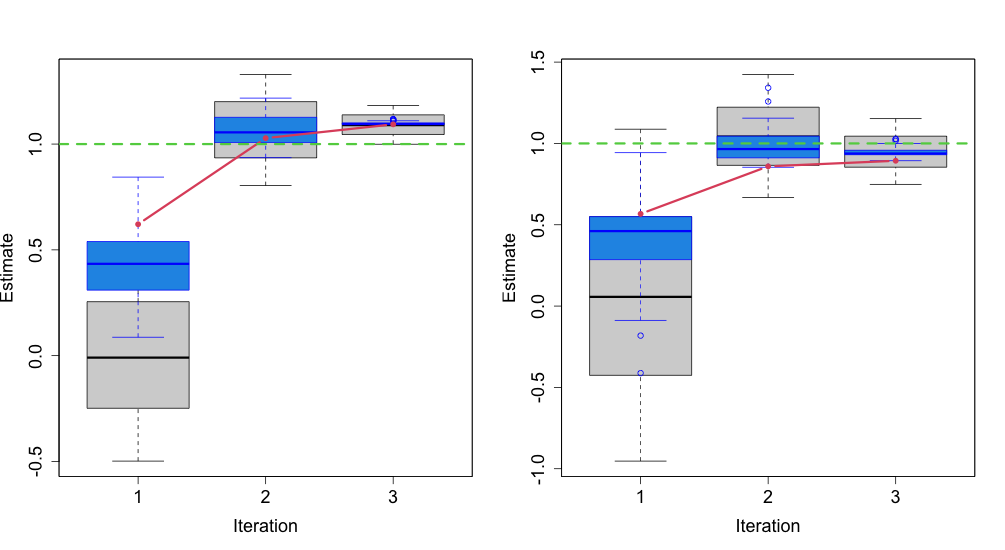}
	\end{subfigure}
	     \begin{subfigure}[b]{0.49\textwidth}
	           			\caption*{\scriptsize{(d) First and second moments}}
		\includegraphics[width=1\textwidth]{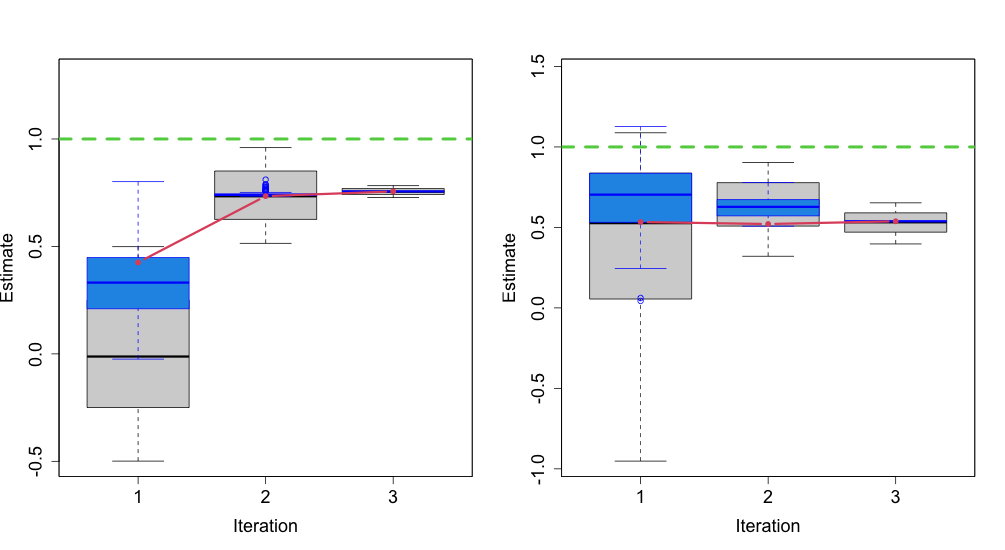}
	\end{subfigure}
    \caption{Boxplots of the training data (grey) and bootstrap uncertainty (blue) at different iterations of Algorithm~\ref{alg2}.
    Points in the red line are the fitted values from the MLP and the green dashed lines are the MLEs.
 Training output data were initially simulated using $N = 10000$ training samples each of length $J=20$ as:
 (a) $\mu_n$ and log($\sigma_n^2$),
 (b) $\mu_n$ and $\sigma_n^2$,
 (c) $\mu_n^2$ and log($\mu_n^2 + \sigma_n^2$), and
 (d) $\mu_n^2$ and $\mu_n^2 + \sigma_n^2$.
 }
		\label{fig:meanvar}
\end{figure}

\paragraph{Results for two independent/dependent parameters} 
We now increase the problem's complexity and evaluate the performance of Algorithm~\ref{alg2} when two parameters are estimated jointly. 
We use the same test data as in the single parameter estimation case (see Figure~\ref{fig:gaussvar}), that is, data from a Gaussian distribution with $\mu_0 = 1$ and $\mbox{log}(\sigma_0^2) =1$.
Specifically, we look at two cases: 1. estimating the mean $\mu_0$ and log-variance $\mbox{log}(\sigma_0^2)$ and 2. estimating the first moment $m_1 = \mu_0$ and the logarithm of the second moment $m_2 = \mbox{log}(\mu_0^2 + \sigma_0^2)$.
Whereas in case 1, the parameters are independent, the MLP has to learn the relation between data and highly dependent parameters in case 2.
In both cases, initial training data does not contain the actual parameters, such that: 1. $\mu_n \sim {\mbox{Unif}}(-0.5, 0.5)$, and
$\mbox{log}(\sigma_{n}^2) \sim {\mbox{Unif}}(-2, 1)$, 2. $m_{1,n} \sim {\mbox{Unif}}(-0.5, 0.5)$ and $m_{2,n} \sim {\mbox{Unif}}\{-0.5^2 + \mbox{exp}(-2), 0.5^2 + \mbox{exp}(1)\}, n=1, \ldots, N$. 
Similarly to the single parameter estimation, we fix $N=10000$ and $J=20$. 
Figure~\ref{fig:meanvar} displays the estimates for Case 1. (top row) and 2. (bottom row) after running Algorithm~\ref{alg2} with the logarithm transformations (left column) and without (right column). 
Estimates are more accurate when the parameters are transformed, and for Case 2, the MLP underestimates the raw moments. 
Especially for the second moment, the algorithm without transformation narrows the estimates close to the median of the initial training data. 
In contrast, the reparametrization is able to recover highly dependent parameters accurately.
Indeed, in both cases, the MLP can detect accurate parameter estimates already in the first iteration.
The subsequent iterations refine the estimates and concentrate the training and the bootstrapping samples around the MLEs. 
This experiment reiterates the benefit of using unbounded and orthogonal parameters for training the MLPs.

\subsection{Spatial extremes}
\label{sec:maxstab}

We now move to a more complex model for spatial extremes, which is well-known to have a likelihood function that is effectively impossible to compute. 
Max-stable distributions are the only possible non-degenerate limits of renormalized pointwise maxima of i.i.d random fields and, therefore, the most commonly used for studying multivariate extreme events \cite{davison2012statistical}. 
We consider the following definition of a max-stable process
\begin{equation}
    X(\mathbf{s}) = \max\limits_{i \geq 1} \xi_i W_i(\mathbf{s}), \mathbf{s} \in \mathcal{S},
    \label{eq:wspec}
\end{equation}
where $\{\xi_i\}_{i\geq 1}$ are points of a Poisson process on $(0,\infty)$ with intensity $d \Lambda (\xi) = \xi^{-2}d\xi$. We consider the Brown-Resnick model \citep{kabluchko2009stationary}, which arises when $W_i(\mathbf{s}) = \mbox{exp}\{ \epsilon_i(\mathbf{s}) - \gamma_i(\mathbf{s})\}$. Each  $\{W_i\}_{i\geq 1}$ is a nonnegative stochastic process with unit mean, whereas $\epsilon_i(\mathbf{s})$ are copies of a zero-mean Gaussian process with semivariogram $\gamma(\mathbf{h}) = (\lVert\mathbf{h}\rVert/\lambda)^\nu$, spatial separation distance $\mathbf{h}$, range $\lambda > 0$, smoothness $\nu \in (0, 2]$ and such that $\sigma^2(\mathbf{h}) = \mbox{var}\{\epsilon(\mathbf{h})\} = 2\gamma(\mathbf{h})$.

The cumulative distribution of $X(\mathbf{s})$ is
\[
p(X(\mathbf{s}_1) \leq x_1, \ldots, X(\mathbf{s}_D) \leq x_D) = \mbox{exp}\{-V( x_1, \ldots,  x_D)\},
\]
where $V( x_1, \ldots,  x_D) = \mbox{E}[\mbox{max}\{W(\mathbf{s}_1)/x_1, \ldots, W(\mathbf{s}_D)/x_D\}]$ satisfies homogeneity and marginal constraints. 
The full likelihood is written as
\[
    f(x_1, \ldots, x_D) =  \mbox{exp}\{-V( x_1, \ldots,  x_D)\} \sum_{\pi \in \Gamma} \prod_{r=1}^{R} \{-V_{\pi_r}(x_1, \ldots,  x_D)\},
\]
where $\Gamma$ is a collection of all partitions $\pi = \{\pi_1, \ldots, \pi_R\}$ of $\{1, \ldots, D\}$ and $V_{\pi_r}$ denotes the partial derivative of $V$ with respect to the variables indexed by $\pi_r$.
The full likelihood is intractable even for moderate $D$ since the number of terms grows equals the \textit{Bell} number, which is more than exponentially. 
The standard workaround for this issue is to consider only pairs of possibly weighted observations in the likelihood \citep{padoan2010likelihood, davis2013statistical, shang2015two}:
\[
  l(\boldsymbol{\theta}) =  \sum_{(j_1, j_2) \in \mathcal{P}} \alpha_{j_1, j_2} \Big[ \mbox{log} \{ V_1(x_{j_1}, x_{j_2}) V_2(x_{j_1}, x_{j_2}) - V_{12}(x_{j_1}, x_{j_2})\} - V_1(x_{j_1}, x_{j_2}) \Big],
\]
where $x_j$ is the block maximum at location $j$, $\Gamma = \{ (j_1, j_2): 1 \leq j_1 < j_2 \leq D \}$, $\boldsymbol{\theta} \in \boldsymbol{\Theta} \subset \mathbb{R}^{P}$ is the vector of unknown parameters and $\alpha_{j_1, j_2} \geq 0$ is the weight of $\{j_1, j_2\}$. 

We compare the estimators from our fully automatic iterative approach with pairwise likelihood estimation on $I=100$ independent simulated datasets of a Brown-Resnick model with $\lambda_0 = 6.2$ and $\nu _0= 1$ on a spatial domain $\mathcal{S}$ of size $[0; 30] \times [0; 30]$ with unit-square grid cells.
The steps used for estimating parameters of Brown-Resnick processes are shown in Algorithm~\ref{alg3}.
To improve accuracy and efficiency, the pairwise likelihood if fit only with pairs with at most 5-units apart and using the \texttt{R}-function \texttt{fitmaxstab} from the \texttt{SpatialExtremes} \texttt{R}-package \citep{ribatet2013spatial}.

    \begin{algorithm}[htb!]
    \caption{Iterative procedure for Brown-Resnick processes}
    \textbf{Need:} Observations $\mathbf{x}_0 \in \mathbb{R}^{J}$ from a distribution $p, \boldsymbol{\theta}_0 \in \mathbb{R}^{P}$ and a neural network $\mathcal{F}_{\phi}(\cdot)$ \\
    Pick $a_{1,p}$ and $a_{2,p}, p=1, 2$, and set $\mathcal{D} = \{\}$ 
    \begin{algorithmic}[1]
    \While{$\mbox{bias}(\hat{\theta}_{0,p}, \tilde{\theta}_{b,p}) > 0.3 \times S_{p}$, for all $p$} 
    \State Sample $\theta_{n,p} \sim \mbox{Unif}(a_{1,p}, a_{2,p}), n = 1, \ldots, N$
    \State Simulate $\mathbf{x}_n \sim p(;\boldsymbol{\theta}_n), n = 1, \ldots, N$    
    \State Set $\mathcal{D}_{\mbox{\scriptsize{train}}} = ({\boldsymbol{\theta}}_n, \mathbf{x}_n)^N_{n=1} $ 
    \State Train $\mathcal{F}_{\phi}(\mathbf{x})$ on  $\mathcal{D}_{\mbox{\scriptsize{train}}} \bigcup \mathcal{D}$ and obtain $\hat{\boldsymbol{\theta}}_0$ from $\mathcal{F}_{\hat{\phi}}(\mathbf{x}_0)$
    \State Simulate $\mathbf{x}_b \sim p(;\hat{\boldsymbol{\theta}}_0)$ and obtain $\hat{\boldsymbol{\theta}}_{b}$ from $\mathcal{F}_{\hat{\phi}}(\mathbf{x}_b), b = 1, \ldots, B$
    \State $a_{1,p} = \hat{\theta}_{0,p} +  \mbox{bias}(\hat{\theta}_{0,p}, \tilde{\theta}_{p}) -  \mathcal{Q}^{0.05}_p(\hat{\theta}_{0,p} - \theta^1_{p}, \ldots, \hat{\theta}_{0,p} -\theta^B_p)$
    \State $a_{2,p} = \hat{\theta}_{0,p} +  \mbox{bias}(\hat{\theta}_{0,p}, \tilde{\theta}_{p}) +  \mathcal{Q}^{0.975}_p(\hat{\theta}_{0,p} - \theta^1_{p}, \ldots, \hat{\theta}_{0,p} -\theta^B_p)$
    \State Randomly select a subset of $\mathcal{D}_{\mbox{\scriptsize{train}}}$ such that $\mathcal{D}_{\mbox{\scriptsize{train}}} \cap \mathcal{D} = \emptyset$ and add those into $\mathcal{D}$
    \EndWhile
   \end{algorithmic}
   \label{alg3}
    \end{algorithm}

 \paragraph{Convolution neural network (CNN)} CNN uses convolutions, that is, the application of a filter to the input image that results in what is called an activation. 
Repeated application of the same filter to images results in a map of activations (feature map). 
This map indicates the locations and strength of a detected feature in the input, such as the edges of objects, and therefore, it is a common choice for regularly-spaced gridded images. 
We use two 2D convolutions with 16 and 8 filters, respectively, and rectified linear unit (ReLU) activation function and kernel of size $3 \times 3$ \citep{hastie2009elements}.
We add one dense layer at the end of the network with four units that map the input image to an output vector of size two. 
The CNN weights are initialized randomly and trained using the Adam optimizer \citep{kingma2014adam} with a learning rate of 0.01.  
The training is performed for 30 epochs, and at each epoch, the CNN weights are updated utilizing a batch size of 100 samples from the entire training dataset.

\paragraph{Initialization} We start the by simulating $N=6000$ pairs $(\boldsymbol{\theta}_n, \mathbf{x}_n)^{N}_{n=1}$, where $\boldsymbol{\theta}_n \equiv (\theta_{n,1}, \theta_{n,2})^{\top} \equiv \{\mbox{log}(\lambda_n), \mbox{logit}(\nu_n)\}^{\top}$ and $\mathbf{x}_n$ is simulated from \eqref{eq:wspec}. 
We initialize $(\theta_{n,1})_{n=1}^N$ based on estimates of a Gaussian process with powered exponential covariance function $C( \mathbf{h}) = \mbox{exp}(-(\lVert \mathbf{h} \rVert/\alpha)^\eta), \alpha>0, 0 < \eta \leq 2$, which closely matches the Brown-Resnick variogram.
We sample $\theta_{n,1} \sim \mbox{Unif}\{\mbox{log}(\hat{\alpha_n}) - c, \mbox{log}(\hat{\alpha_n}) + c \}$.
Such estimates are likely biased for Brown-Resnick, but they are quick to compute and a better start than a random guess.
A line search for $c$ indicates that $c=2$ provides good results but found that other values of $c$ gave similar accuracy.
We uniformly draw $(\theta_{n,2})_{n=1}^{N}$ over approximately the whole (bounded) domain:
$\theta_{n,2} \sim \mbox{Unif}\{\mbox{logit}(0.1), \mbox{logit}(1.9) \}$.
We initialize the pairwise likelihood with the powered exponential covariance function estimates for a fair comparison with our approach.

If the parameter region for simulating data at each iteration is tightened too quickly or shifted by too much, it may miss the true parameters.
To alleviate this issue, we randomly select $40\%$ of the samples from the previous iteration that were not contained in the updated interval and add them to the training data of the current step (see line 9 of Algorithm~\ref{alg3}).  
Besides increasing the size of the training without having to simulate new data, this broadens the range of the training while still keeping most of the samples in the updated region defined by the current step. 
Therefore, it prevents the current iteration from being stuck in the wrong region while maintaining more accurate training data in the most probable parameter region.

\begin{figure}[htb!]
	\centering
		\includegraphics[width=0.5\textwidth]{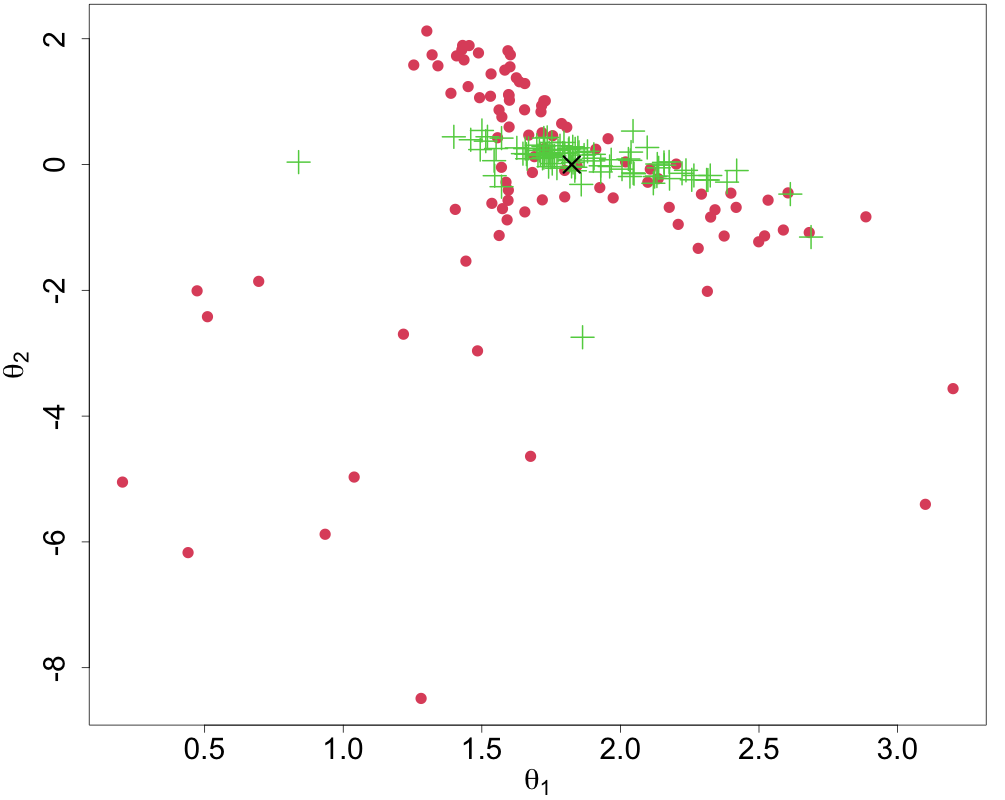}
	\caption{Scatterplots of estimated parameters on the transformed scales. Each dot/cross shows 100 independent estimates from the Brown-Resnick model using the CNN (green) or PL (red). 
	The $\times$ is the actual value.
	 Training datasets were initially simulated using $N = 6000$ training samples on a $[0,30]^2$ and based on estimates from fitting a powered exponential covariance function to the data.}
	\label{fig:scater_br}
\end{figure}

\begin{figure}[htb!]
	\centering
		\includegraphics[width=0.75\textwidth]{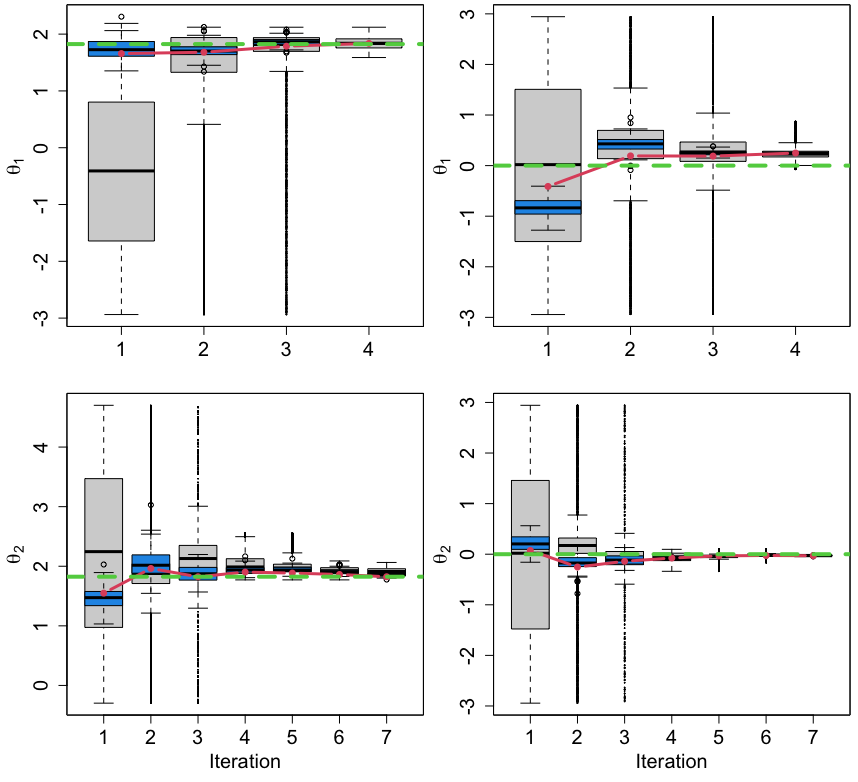}
	\caption{Boxplots of the training data (grey) and bootstrap uncertainty (blue) at different iterations of Algorithm~\ref{alg3} for $\theta_1$ (left) and $\theta_2$ (right). 
The top/bottom row shows an example of when the variogram estimates are outside/inside the initial training data. 
Points in the red line are the fitted values from the CNN, and the green dashed lines are the truth.}
	\label{fig:box_br}
\end{figure}

Figure~\ref{fig:scater_br} shows a scatterplot of 100 independent estimates of $\theta_{1} \equiv \mbox{log}(\lambda_0)$ versus $\theta_{2} \equiv \mbox{log}(\nu_0)$  from the last iteration of our approach (green) and from the pairwise likelihood (red). 
The $\times$ symbol is the truth. 
Whereas the proposed method produces robust results across the different replicates, the pairwise likelihood tends to underestimate the smoothness parameter, and the performance varies considerably across datasets.
In Figure~\ref{fig:box_br}, we access the accuracy of Algorithm~\ref{alg3} for estimating $\theta_{1}$ (left column) and $\theta_{2}$ (right column) with boxplots.
The rows in this figure illustrate the results for two different datasets: 
The first is initialized with training data that do not contain $\theta_1$ (top), whereas the variogram estimate for the second dataset is close to the center of the training data. 
Indeed, whereas the space covering $\theta_2$ is bounded, and simulating training data covering the entire region is straightforward, the training data for $\theta_2$ is based on the variogram estimate and, therefore, only sometimes contains the truth. 
The grey and blue boxplots at each plot and iteration are the training output and bootstrapping samples, respectively.
As expected, the gray boxplots in all cases at iterations 2 and 3 contain several outliers, which correspond to the samples reused from the previous step.
Points in the red line are the fitted values from the CNN, and the green dashed lines are the actual parameters used to simulate data.

Even when the true value is not included in the initial training data (see the top of Figure~\ref{fig:box_br}), the CNN estimates $\theta_1$ well and produces reasonable uncertainties. 
At the last iteration, the training data for both parameters are narrow and around the actual value, and the bootstrapping samples practically coincide with the training data. 
When initialized with training data containing the truth, the CNN quickly approaches the truth with low uncertainty for both parameters and remains stable until it reaches the stopping criteria.
A quantitative measure of the effect of the iterations in Algorithm~\ref{alg3} is reported in Table~\ref{tab:iter-errors}, with bias, standard deviation, and root mean square error (RMSE) for the first and last iterations.
The metrics are calculated from the bootstrapping samples among the 100 independent datasets. 
Point estimates are taken as the median of the bootstrapping samples. 
Under all three metrics, there is a considerable improvement from the first to the last iteration of the algorithm, and estimation at the last iteration are about $15\%$ more efficient for $\theta_1$ and $30\%$  more efficient for $\theta_2$ (with the efficiency defined as the ratio of RMSEs).

\begin{table}[htb!]
\centering
  \begin{tabular}{l|l|l|l|l|l|l|}
      & \multicolumn{3}{c|}{$\theta_1$} &
      \multicolumn{3}{c|}{$\theta_2$} \\
    Iteration & bias & sd & RMSE & bias & sd & RMSE\\
          \hline
    First & 0.223 & 0.295 & 0.563 & 0.005 & 0.253 & 0.451 \\
    \hline
    Last & 0.056 & 0.129 & 0.306 & 0.004 & 0.120 & 0.313 \\
  \end{tabular}
  \caption{Standard deviation, bias and RMSE of the estimated parameters from the first and last iteration of Algorithm~\ref{alg3}}
\label{tab:iter-errors}
\end{table}

\section{A general unified database approach for time series}
\label{sec:ts}

\subsection{General framework}

Suppose we observe time series data $\mathbf{x}_0 = \{x_0(1), \ldots, x_0(T) \}^{\top}$ from a strictly stationary process  $\{X_0(t): t \in \mathcal{T}\}$ indexed on the temporal domain $T \subset \mathbb{R}_{+}$. 
Let $p(·; \boldsymbol{\theta}_0)$ be the probability distribution of $X_0(t)$ depending on the parameter set $\boldsymbol{\theta}_0 \in \Theta \subset \mathbb{R}^P$. 
The stationarity assumption is that the joint probability distribution of $\{X_0(t-l), \ldots, X_0(t)\}^{\top}$ does not depend on $X_0(t-l')$ for any $l'>l$.
This Markov property is common in time series analysis similarly with ergodicity, which provides the justification for estimating $\boldsymbol{\theta}_0$ from a single sequence $\mathbf{x}_0$.

\begin{figure}[htb!]
\centering
\includegraphics[width=0.7\textwidth]{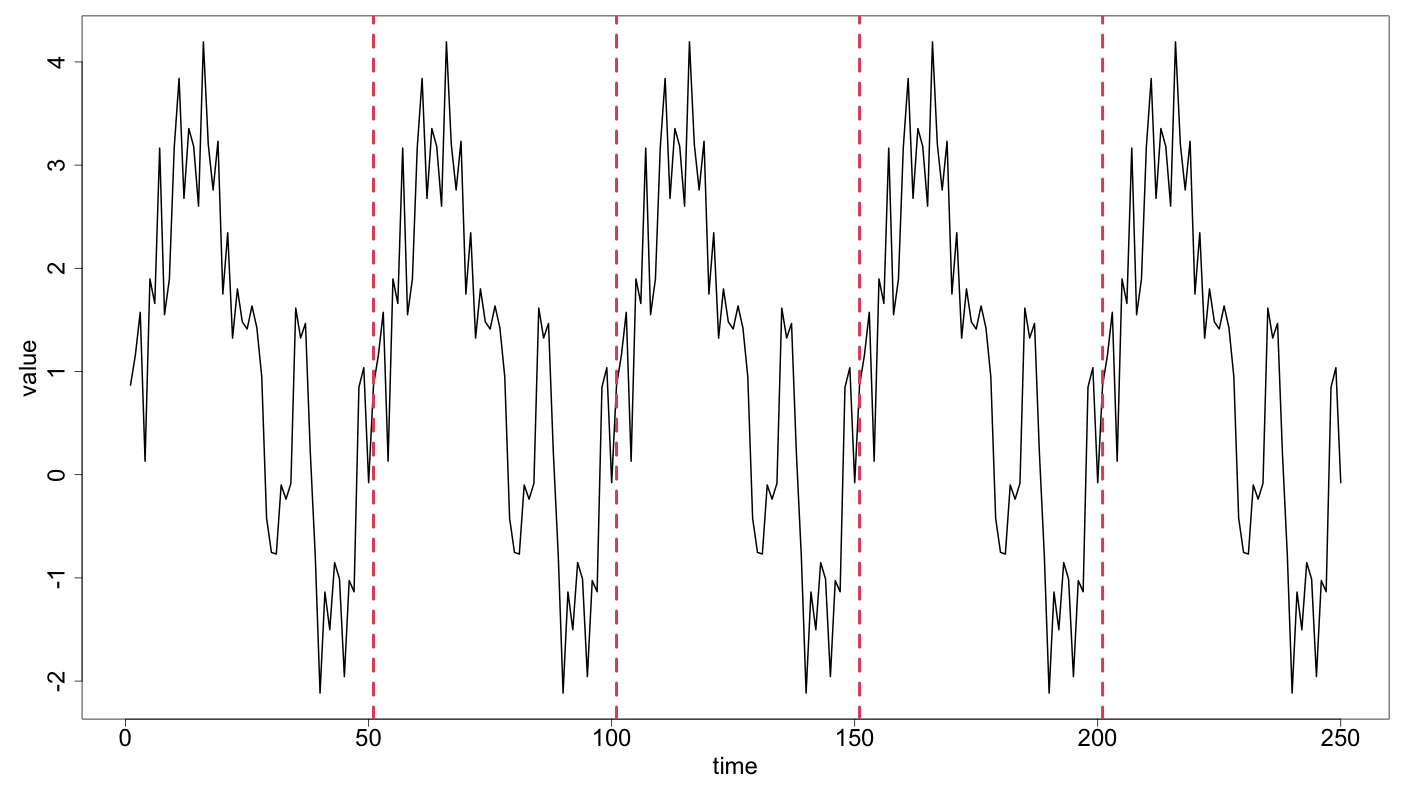}
	\caption{Data from an autoregressive process of order one (AR(1)) of length $T=50$ with AR coefficient equal to 0.9, augmented five times to achieve the training data length $T_k = 250$.}
\label{fig:eg-ts}
\end{figure}

Our main goal is to estimate $\boldsymbol{\theta}_0$ by training a DNN $\mathcal{F}_{\phi}$ using parameter candidates as output and corresponding simulated data as input (see Section~\ref{sec:parm-est}). 
Here, we take advantage of the stationarity property to generalize and improve the estimation workflow described in Section~\ref{sec:parm-est}. 
Our approach is best exemplified by a toy data  $\mathbf{x}_0 = \{x_0(1), \ldots, x_0(T) \}^{\top}$ with $T=50$ from an AR process of order 1 with coefficient $\rho_1 = 0.9$. 
Instead of simulating training data of length $T=50$, we proceed by simulating time series of length $T_k = 250$ and construct data pairs $(\boldsymbol{\theta}_n, \mathbf{x}_n)^{N}_{n=1}$, where $\boldsymbol{\theta}_n \in \mathbb{R}^P$ and $\mathbf{x}_n = \{x_n(1), \ldots, x_n(250) \}^{\top}$ and train a DNN.
Next, since $\mathbf{x}_0$ is shorter than the training data, we create a new time series $\mathbf{x}^{*}_0$ by concatenating $\mathbf{x}_0$ to achieve the length of the time series used during training.
Figure~\ref{fig:eg-ts} shows how to construct $\mathbf{x}^*_0$ from $\mathbf{x}_0$ by replicating the observations five times. 
The red dashed line are the joining points.
The resulting $\mathbf{x}^{*}_0$ is then fed into the trained DNN to retrieve estimates $\hat{\boldsymbol{\theta}}_0$.

Recurrent neural networks take information from prior inputs to predict the current input and output and can handle sequence data and inputs of varying lengths by memorizing historical information. 
Therefore, these networks are unsuitable for our problem, as our output data does not have future sequence values. 
Here, we are interested in mapping parameters to sequential data rather than learning the next value in a sequence.
As we will see in the example in the next section, our technique has several advantages over training the DNN using simulated data with the same length as the observations: (i) allows estimation of time series of several sizes at almost no computational cost, since the DNN does not have to be retrained for each new dataset, 
(ii) holds without requiring any particular dependence structure assumption as long as the data is stationary and Markov.

\paragraph{Connection to non-overlapping block bootstrap (NBB)} The intuition behind our approach resembles NBB approaches \citep{carlstein1986use}. Similarly, this technique splits the observations into non-overlapping blocks and resamples the blocks with replacement, which are replicated to obtain a bootstrapped series.
However, unlike block bootstrap methods, where the complex problem of choosing the block size has to be solved, by construction, our block size is always fixed and equal to $T$.

\paragraph{Discontinuity at the joining points} Our procedure of laying sequences of length $T$ end-to-end will inevitably produce $m - 1$ discontinuity points where the joining occurs, similarly to what happens in bootstrap for time series. 
However, as we will show empirically in our examples, these discontinuities will have a negligible contribution to the model parameters structure.

\paragraph{Varying observation lengths} 
As long as the database for training the neural network is extensive enough, the proposed method is easily generalized for cases where $T$ is not a multiple of $T_k$.
If the observed time series is shorter than that used during training, that is $T < T_k$,  one can replicate the data into $m$ blocks, where $m = \lfloor \frac{T_k}{T} \rfloor$, and complete the remaining values with a random block from $\mathbf{x}_0$ of size $T_k \mbox{mod} T$.
The idea is that if $T_k$ is large enough compared to $T$ and with  $mT  \backsimeq T_k$, as $m \rightarrow \infty$ and $T_k \rightarrow \infty$, the last components of $\mathbf{x}^*_0$ have little influence in the dependence structure.
On the other hand, if $T > T_k$, we take the first $T_k$ observations and estimate parameters $\boldsymbol{\theta}^{(1)}_0$, the next $T_k$ and estimate $\boldsymbol{\theta}^{(2)}_0$, and so on. 
If the last sequence from this procedure is smaller than $T_k$, we use the method for dealing with $T < T_k$ described previously. 
Next, we merge the results from these several estimates (e.g., by using the sample average) to create a single combined estimate $\hat{\boldsymbol{\theta}}_0$, which can then be used to simulate bootstrap samples and construct confidence intervals.

\paragraph{Uncertainty quantification} 
To account for the underestimation in the uncertainty that results from applying the estimator to the replicated time series, we scale the sampling variance by the number of blocks used to replicate the original series. 
For instance, consider the simulated toy example in Figure~\ref{fig:eg-ts} of an AR(1) process with $T=200$ and an AR coefficient $\rho = 0.9$. 
The MLE is $\hat{\rho}_{T} \approx 0.8320$, and its corresponding standard deviation is $0.0784$. 
If the series is then replicated $5$ times such that the new length is $T_k=250$, the new MLE is not affected ($\hat{\rho}_{T_k} \approx 0.8320$), but its standard deviation is now $0.0357$.  
Therefore, to recover the correct uncertainty, we need to scale the new standard deviation by $\sqrt{5}$, that is, $0.0357 \times \sqrt{5} \approx 0.0798$, which is then, approximately, the standard deviation of $\hat{\rho}_{T}$.
This adjusts the uncertainty for the fact that no additional information is added to the replicated data.

\paragraph{1D Convolutional Neural Networks} 
Since we are dealing with parameters from time series data, we train 1D CNNs, which have proven successful in learning features from dependent observations onto one-dimensional dependent sequences.
As for 2D CNNs (see Section~\ref{sec:maxstab} for an example), the input layer receives the (transformed) data, and the output layer is an MLP with the number of neurons equal to the number of output variables. 
Each neuron in a hidden layer first performs a sequence of convolutions, the sum of which is passed through the activation function followed by a sub-sampling operation. 
The early convolutional layers can be seen as smoothing the input vector, where the filters are similar to parameters of a weighted moving average but learned jointly with the regression parameters from the MLP layer.
In what follows, we use three 1D convolutions with four filters each, the ReLU activation function, a kernel of size three, and one final dense layer with four units.
We set a learning rate of 0.01 with 30 epochs and a batch size of 50 samples to update the weights. 

\subsection{Non-Gaussian stochastic volatility model}
To show the usefulness of our approach, we focus on estimating parameters of financial time series data that exhibit non-Gaussian time-varying volatility.
Volatility is highly right-skewed and bounded, making Gaussian distributions a poor representation.
A better description of volatility is achieved with stochastic volatility models (SVOL), first introduced in \citet{taylor1982financial} and currently central to econometrics and finance investments theory and practice. 
The idea of SVOL models is to parsimoniously fit the volatility process as a latent structure using an unconditional approach that does not depend on observations.
We consider the following model structure
\begin{equation}
\begin{aligned}
   x(t) & = \sqrt{\mbox{exp}\{h(t)\}} \epsilon_t, \quad \sqrt{\frac{\nu - 2}{\nu}} \epsilon(t)\sim T_{\nu}, \quad t=1, \ldots, T \\
    h(t) & = \rho h(t-1) + \xi_t,  \quad \xi(t)\sim N(0, \sigma^2),
\end{aligned}
    \label{eq:sv}
\end{equation}
where $\epsilon(t)$ and $\xi(t)$ are independent noises and $T$ is the number of observations. The volatility variable $h(t)$ is latent with an AR(1) structure, and only $x(t)$ is observed. 
When $\lvert{\rho}\rvert < 1$, $x(t)$ is strictly stationary with mean $\mu_{h} = 1/(1 - \rho)$ and variance $\sigma^2_{h} = \sigma^2/(1 - \rho^2)$ \citep{fridman1998maximum}.

Likelihood evaluation of continuous dynamic latent-variable models such as \eqref{eq:sv} requires the integration of the latent process out of the joint density, resulting in the following $T$-dimensional integral 
\begin{equation}
\begin{aligned}
l({\boldsymbol{\theta}}) & = \int_{\mathbf{h}} p(\mathbf{x}\mid \mathbf{h}, {\boldsymbol{\theta}}) p(\mathbf{h} \mid {\boldsymbol{\theta}}) d\mathbf{h} \\
 & = \int_{\mathbf{h}} \prod^T_{t=1} p(\mathbf{x}\mid \mathbf{h}) p\{h(t)\mid h(t-1)\} d\mathbf{h},
\end{aligned}
 \label{eq:ml-sv}
\end{equation}
where $\mathbf{x} = \{x(1), \ldots, x(T)\}^{\top}$, $\mathbf{h} = \{h(1), \ldots, h(T)\}^{\top}$, the conditional densities $p(\mathbf{x}\mid \mathbf{h}, {\boldsymbol{\theta}})$ and $ p\{h(t)\mid h(t-1)\}$ have the form in \eqref{eq:sv}, and
the initial volatility $h(0)$ is the stationary volatility distribution $p(\mathbf{h})$. 
Since $h(t)$ is not independent from the past, the integral in \eqref{eq:ml-sv} cannot be factored into a product of $T$ one-dimensional integrals and exact evaluation of the likelihood is possible only in special cases like when both $p(\mathbf{x}\mid \mathbf{h}, {\boldsymbol{\theta}})$ and $p\{h(t)\mid h(t-1)\}$ are Gaussian. 
Alternative approaches for likelihood evaluation include computationally demanding Markov Chain Monte Carlo (MCMC) \citep{andersen1999efficient} and the more recent and faster Integrated Nested Laplace Approximation (INLA) \citep{martino2011estimating}. 

Next, we give practical details of our framework as well as the a comparison of the results from our approach and the state-of-the-art INLA approach for estimating parameters of the SVOL model.

\paragraph{Implementation}
Consider observations $\mathbf{x}_{0} = \{x_{0}(1), \ldots, x_{0}(T) \}^{\top}$, from the SVOL model  with $\sigma_0=0.1$, $\rho_0 = 0.8$ and $\nu_0 = 6$.
We use scaled (variance one) versions of both $h(t)$ and $x(t)$ in  \eqref{eq:sv}, such that only $\rho_0$ and $\nu_0$ need to be estimated.
To show the effect of estimating time series of different lengths with a single DNN fit, we display the results for various time series lengths: $T = (500, 1000, 2000, 3000, 4000, 5000)$.
Estimation goes as follows. 
The training database contain $N=10000$ samples pairs $(\boldsymbol{\theta}_{n}, \mathbf{x}_{n})^{10000}_{n=1}$ of transformed parameters $\boldsymbol{\theta}_{n} = (\theta_{1}, \theta_{2})^{\top} = \{f_1(\rho_{n}), f_2(\nu_{n})\}^{\top}$, with $f_1(x) = \mbox{log} \Big( \frac{1+ x}{1 - x}\Big)$ and $f_2(x) = \mbox{log}(x - 2)$ 
and corresponding data $\mathbf{x}_n = \{x_n(1), \ldots, x_n(5000) \}^{\top}$ simulated from \eqref{eq:sv}.
The transformed parameters are sampled uniformly in a neighborhood of the actual parameter values $\theta_{0,1} = f_1(\rho_0)$ and $\theta_{0,2} = f_2(\nu_0)$:
\begin{equation}
    \begin{aligned}
    f_1(\rho_j) & \sim {\mbox{Unif}}\{\theta_{0,1} - c^a_1, \theta_{0,1} + c^b_1\} \\
    f_2(\nu_j)  & \sim {\mbox{Unif}}(\theta_{0,2} - c^a_2, \theta_{0,2} + c^b_2). \\
    \label{eq:ts_sim}
\end{aligned}
\end{equation}
New test data $\mathbf{x}^*_{0}$ is obtained by replicating $\mathbf{x}_{0}$  $m$ times as many time as needed to achieve size $5000$.
We fix $c^a_1 = c^a_2 = c^b_1 = c^b_2 = 2$ in \eqref{eq:ts_sim} to ensure a large enough region around the true values, although other constants provided similar results.

\begin{figure}[htb!]
\centering
\includegraphics[width=1\textwidth]{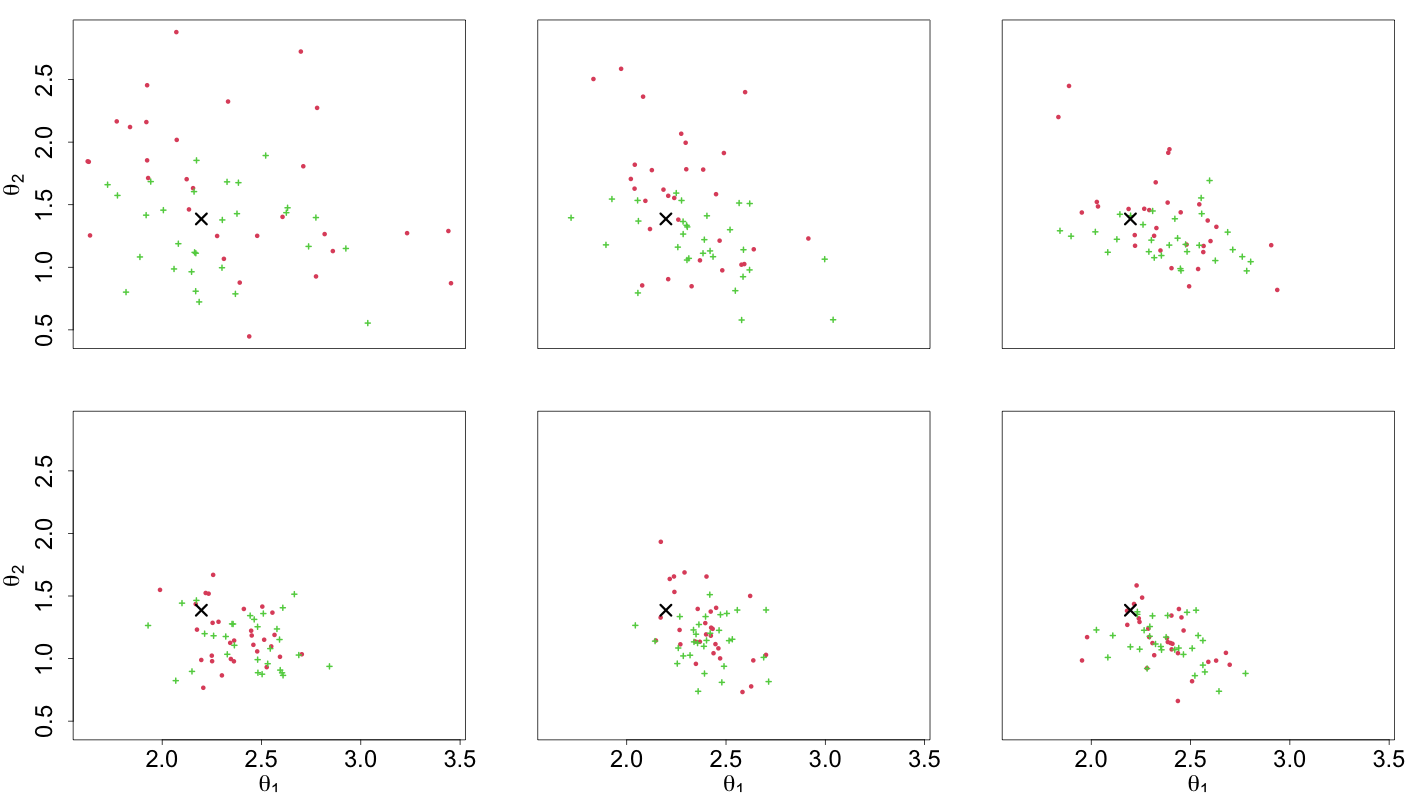}
	\caption{Scatterplots of estimated parameters on the transformed scales. Each plot shows 30 independent estimates from the SVOL model from the 1D CNN (green) and INLA (red). Small testing data to large are displayed from top to bottom, left to right:  $T = \{500, 1000, 2000, 3000, 4000, 5000\}$. The $\times$'s are the true values.
	}
\label{fig:hist-cnn-inla}
\end{figure}

Figure~\ref{fig:hist-cnn-inla} displays scatterplots of estimated $\theta_{0,1}$ versus estimated $\theta_{0,2}$ from $I=30$ independent replicates of the SVOL model. 
Scatterplots from top to bottom, left to right, shows testing sets of different size: $T = (500, 1000, 2000, 3000, 4000, 5000)$.
Green dots are the 1D CNN estimates, and red dots are the mean of the predicted posterior distribution from fitting model \eqref{eq:sv} using INLA.
The $\times$ symbol in each plot represents the truth. 
As the testing data sizes increase, both methods concentrate the estimates around the truth, and the 1D CNN estimates are less variable and less biased for smaller values of $T$; after that, both methods seem to perform similarly well.
Whereas we set the INLA priors to default values, changing them to penalize parameter values far from the mode could potentially improve the results.

\begin{figure}[htb!]
\centering
				\begin{subfigure}[b]{1\textwidth}
\includegraphics[width=0.5\textwidth]{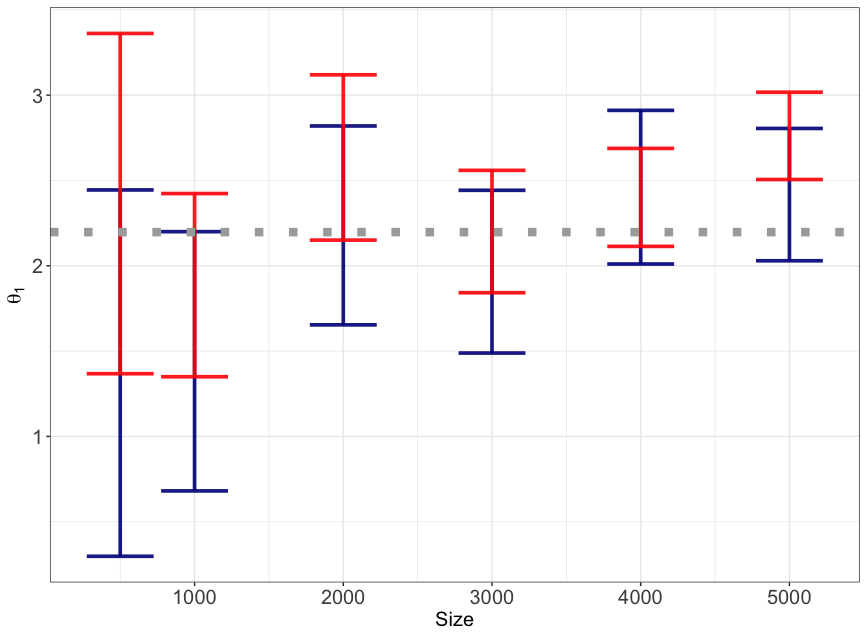}
\includegraphics[width=0.5\textwidth]{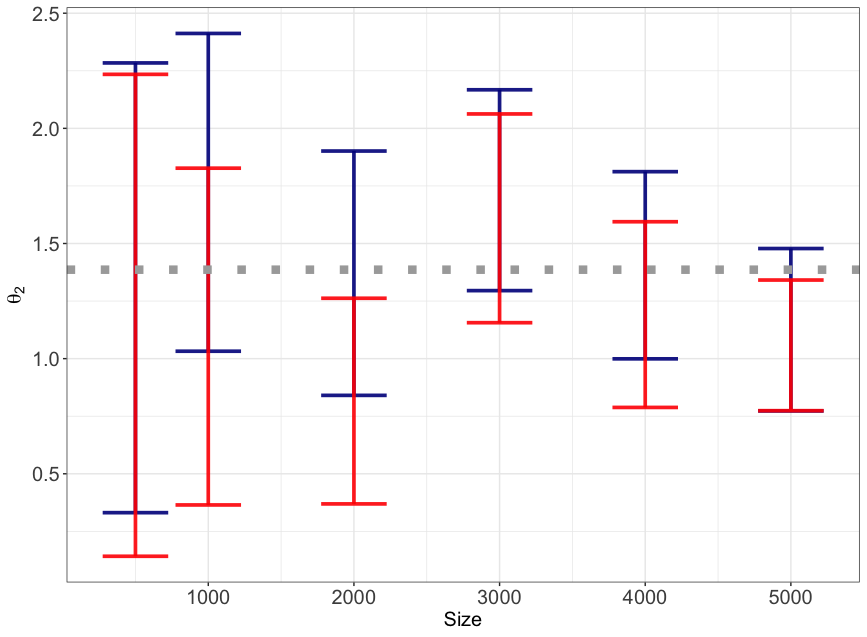}
\end{subfigure}
	\caption{Example of $95\%$ estimated central interval for $\theta_{0,1}$ (left) and $\theta_{0,2}$ (right) using the estimated posterior distribution from INLA (red) and 10000 bootstrapped samples from the 1D CNN (blue). 
	The $x$-axis are the different test data sizes $T = (500, 1000, 2000, 3000, 4000, 5000)$ and the horizontal dashed line is the truth.}
\label{fig:ic-cnn-inla}
\end{figure}

To quantify the uncertainty in the 1D CNN estimates, we create a bootstrapped dataset by independently sampling $B=100000$ time series from the fitted model and then feeding these samples into the trained 1D CNN. 
The uncertainty in the estimation of $\theta_{0,1}$ (left) and $\theta_{0,2}$ (right) from both methods is grasped in Figure~\ref{fig:ic-cnn-inla}. 
The bars represent central $95\%$ intervals, taken from the posterior distribution given by INLA (red) and the 1D CNN bootstrapped samples (blue) for one randomly chosen dataset among the $I=30$ replicates in Figure~\ref{fig:hist-cnn-inla}.
The $x$-axis represents different test data sizes $T = (500, 1000, 2000, 3000, 4000, 5000)$, and the gray horizontal dashed line is the truth.
As expected, the uncertainties decrease with sample sizes for both methods. 
The intervals from both methods are close for most data sizes, although the INLA distributions are more concentrated for $\theta_{0,1}$ and $T=(4000, 5000)$.
Overall, these results show that the 1D CNN is robust to estimating parameters of different data lengths and in agreement with the INLA estimator.


\section{Conclusion}
\label{sec:conclusion}

We proposed approaches that train DNNs to estimate parameters of intractable models and quantify their uncertainty.
Unlike previously proposed approaches using DNNs, which are tailored to a specific application and can lead to poor parameter estimates for relying on computationally expensive initial guesses to construct training data, our methods 
(A) leverage an iterative learning framework coupled with a modified parametric bootstrap step to guide simulations in the direction of the parameter region of the actual data in multiple rounds
(B) use an extensive pre-trained database to accurately estimate parameters of time series data of multiple lengths at no computational cost, rather than simulating data for every new dataset.

Our estimators yield accurate parameter estimates with much less computation time than classical methods, even when accounting for the time required to generate training samples.
The experiments discussed in this work involve models with a maximum of two parameters for clarity in presentation. 
We anticipate a more pronounced influence of our sequential approach in higher-dimensional scenarios for alleviating the challenges associated with the ``curse of dimensionality" as the volume of the parameter space increases exponentially with the number of parameters.

While DNNs for parameter estimation are gaining popularity, we still need to learn more about black-box algorithms applied to previously intractable statistical problems and how to design task-specific estimators more generally. 
There are several further opportunities for exploring DNNs for parameter estimation using newly designed optimization tools from the machine learning community.
This work is another step towards this direction, where ultimately, inference is performed within a general and flexible simulation-based black-box pipeline.

\baselineskip=18.5pt

\bibliography{sample}

\begin{thebibliography}{}

\bibitem[Andersen et~al., 1999]{andersen1999efficient}
Andersen, T.~G., Chung, H.-J., and S{\o}rensen, B.~E. (1999).
\newblock Efficient method of moments estimation of a stochastic volatility
  model: A monte carlo study.
\newblock {\em Journal of econometrics}, 91(1):61--87.

\bibitem[Beaumont et~al., 2009]{beaumont2009adaptive}
Beaumont, M.~A., Cornuet, J.-M., Marin, J.-M., and Robert, C.~P. (2009).
\newblock Adaptive approximate bayesian computation.
\newblock {\em Biometrika}, 96(4):983--990.

\bibitem[Bonassi and West, 2015]{bonassi2015sequential}
Bonassi, F.~V. and West, M. (2015).
\newblock Sequential monte carlo with adaptive weights for approximate bayesian
  computation.

\bibitem[Carlstein, 1986]{carlstein1986use}
Carlstein, E. (1986).
\newblock The use of subseries values for estimating the variance of a general
  statistic from a stationary sequence.
\newblock {\em The annals of statistics}, pages 1171--1179.

\bibitem[Cranmer et~al., 2020]{cranmer2020frontier}
Cranmer, K., Brehmer, J., and Louppe, G. (2020).
\newblock The frontier of simulation-based inference.
\newblock {\em Proceedings of the National Academy of Sciences},
  117(48):30055--30062.

\bibitem[Davis et~al., 2013]{davis2013statistical}
Davis, R.~A., Kl{\"u}ppelberg, C., and Steinkohl, C. (2013).
\newblock Statistical inference for max-stable processes in space and time.
\newblock {\em Journal of the Royal Statistical Society: SERIES B: Statistical
  Methodology}, pages 791--819.

\bibitem[Davison et~al., 2012]{davison2012statistical}
Davison, A.~C., Padoan, S.~A., and Ribatet, M. (2012).
\newblock Statistical modeling of spatial extremes.
\newblock {\em Statistical science}, 27(2):161--186.

\bibitem[Drovandi and Frazier, 2022]{drovandi2022comparison}
Drovandi, C. and Frazier, D.~T. (2022).
\newblock A comparison of likelihood-free methods with and without summary
  statistics.
\newblock {\em Statistics and Computing}, 32(3):42.

\bibitem[Fearnhead and Prangle, 2012]{fearnhead2012constructing}
Fearnhead, P. and Prangle, D. (2012).
\newblock Constructing summary statistics for approximate bayesian computation:
  semi-automatic approximate bayesian computation.
\newblock {\em Journal of the Royal Statistical Society: Series B (Statistical
  Methodology)}, 74(3):419--474.

\bibitem[Frazier et~al., 2018]{frazier2018asymptotic}
Frazier, D.~T., Martin, G.~M., Robert, C.~P., and Rousseau, J. (2018).
\newblock Asymptotic properties of approximate bayesian computation.
\newblock {\em Biometrika}, 105(3):593--607.

\bibitem[Fridman and Harris, 1998]{fridman1998maximum}
Fridman, M. and Harris, L. (1998).
\newblock A maximum likelihood approach for non-gaussian stochastic volatility
  models.
\newblock {\em Journal of Business \& Economic Statistics}, 16(3):284--291.

\bibitem[Friedman et~al., 2001]{friedman2001elements}
Friedman, J., Hastie, T., Tibshirani, R., et~al. (2001).
\newblock {\em The {E}lements of {S}tatistical {L}earning}, volume~1.
\newblock Springer Series in Statistics.

\bibitem[Gerber and Nychka, 2020]{gerber2021fast}
Gerber, F. and Nychka, D.~W. (2020).
\newblock Fast covariance parameter estimation of spatial {G}aussian process
  models using neural networks.
\newblock {\em Stat}, page e382.

\bibitem[Gourieroux et~al., 1993]{gourieroux1993indirect}
Gourieroux, C., Monfort, A., and Renault, E. (1993).
\newblock Indirect inference.
\newblock {\em Journal of applied econometrics}, 8(S1):S85--S118.

\bibitem[Grelaud et~al., 2009]{grelaud2009abc}
Grelaud, A., Marin, J.-M., Robert, C.~P., Rodolphe, F., and Taly, J.-F. (2009).
\newblock Abc likelihood-free methods for model choice in gibbs random fields.
\newblock {\em Bayesian Analysis}, 4(2):317--335.

\bibitem[Gutmann and Corander, 2016]{gutmann2016bayesian}
Gutmann, M.~U. and Corander, J. (2016).
\newblock Bayesian optimization for likelihood-free inference of
  simulator-based statistical models.
\newblock {\em Journal of Machine Learning Research}.

\bibitem[Hartig et~al., 2011]{hartig2011statistical}
Hartig, F., Calabrese, J.~M., Reineking, B., Wiegand, T., and Huth, A. (2011).
\newblock Statistical inference for stochastic simulation models--theory and
  application.
\newblock {\em Ecology letters}, 14(8):816--827.

\bibitem[Hastie et~al., 2009]{hastie2009elements}
Hastie, T., Tibshirani, R., and Friedman, J. (2009).
\newblock {\em The Elements of Statistical Learning: Data Mining, Inference,
  and Prediction}.
\newblock Springer Science \& Business Media.

\bibitem[J{\"a}rvenp{\"a}{\"a} et~al., 2019]{jarvenpaa2019efficient}
J{\"a}rvenp{\"a}{\"a}, M., Gutmann, M.~U., Pleska, A., Vehtari, A., and
  Marttinen, P. (2019).
\newblock Efficient acquisition rules for model-based approximate bayesian
  computation.

\bibitem[Jiang et~al., 2017]{jiang2017learning}
Jiang, B., Wu, T.-y., Zheng, C., and Wong, W.~H. (2017).
\newblock Learning summary statistic for approximate {B}ayesian computation via
  deep neural network.
\newblock {\em Statistica Sinica}, pages 1595--1618.

\bibitem[Kabluchko et~al., 2009]{kabluchko2009stationary}
Kabluchko, Z., Schlather, M., De~Haan, L., et~al. (2009).
\newblock Stationary max-stable fields associated to negative definite
  functions.
\newblock {\em The Annals of Probability}, 37(5):2042--2065.

\bibitem[Kingma and Ba, 2014]{kingma2014adam}
Kingma, D.~P. and Ba, J. (2014).
\newblock Adam: A method for stochastic optimization.
\newblock {\em arXiv preprint arXiv:1412.6980}.

\bibitem[Lenzi et~al., 2021]{lenzi2021neural}
Lenzi, A., Bessac, J., Rudi, J., and Stein, M.~L. (2021).
\newblock Neural networks for parameter estimation in intractable models.
\newblock {\em arXiv preprint arXiv:2107.14346}.

\bibitem[Lueckmann et~al., 2019]{lueckmann2019likelihood}
Lueckmann, J.-M., Bassetto, G., Karaletsos, T., and Macke, J.~H. (2019).
\newblock Likelihood-free inference with emulator networks.
\newblock In {\em Symposium on Advances in Approximate Bayesian Inference},
  pages 32--53. PMLR.

\bibitem[Lueckmann et~al., 2017]{lueckmann2017flexible}
Lueckmann, J.-M., Goncalves, P.~J., Bassetto, G., {\"O}cal, K., Nonnenmacher,
  M., and Macke, J.~H. (2017).
\newblock Flexible statistical inference for mechanistic models of neural
  dynamics.
\newblock {\em Advances in neural information processing systems}, 30.

\bibitem[Martino et~al., 2011]{martino2011estimating}
Martino, S., Aas, K., Lindqvist, O., Neef, L.~R., and Rue, H. (2011).
\newblock Estimating stochastic volatility models using integrated nested
  laplace approximations.
\newblock {\em The European Journal of Finance}, 17(7):487--503.

\bibitem[Murphey et~al., 2004]{murphey2004neural}
Murphey, Y.~L., Guo, H., and Feldkamp, L.~A. (2004).
\newblock Neural learning from unbalanced data.
\newblock {\em Applied Intelligence}, 21(2):117--128.

\bibitem[Nickl and P{\"o}tscher, 2010]{nickl2010efficient}
Nickl, R. and P{\"o}tscher, B.~M. (2010).
\newblock Efficient simulation-based minimum distance estimation and indirect
  inference.
\newblock {\em Mathematical methods of statistics}, 19(4):327--364.

\bibitem[Padoan et~al., 2010]{padoan2010likelihood}
Padoan, S.~A., Ribatet, M., and Sisson, S.~A. (2010).
\newblock Likelihood-based inference for max-stable processes.
\newblock {\em Journal of the American Statistical Association},
  105(489):263--277.

\bibitem[Papamakarios and Murray, 2016]{papamakarios2016fast}
Papamakarios, G. and Murray, I. (2016).
\newblock Fast $\varepsilon$-free inference of simulation models with bayesian
  conditional density estimation.
\newblock In {\em Advances in neural information processing systems}, pages
  1028--1036.

\bibitem[Papamakarios et~al., 2019]{papamakarios2019sequential}
Papamakarios, G., Sterratt, D., and Murray, I. (2019).
\newblock Sequential neural likelihood: Fast likelihood-free inference with
  autoregressive flows.
\newblock In {\em The 22nd International Conference on Artificial Intelligence
  and Statistics}, pages 837--848. PMLR.

\bibitem[Ribatet, 2013]{ribatet2013spatial}
Ribatet, M. (2013).
\newblock Spatial extremes: Max-stable processes at work.
\newblock {\em Journal de la Soci{\'e}t{\'e} Fran{\c{c}}aise de Statistique},
  154(2):156--177.

\bibitem[Richards et~al., 2023]{jordan2023neural}
Richards, J., Sainsbury-Dale, M., Huser, R., and Zammit-Mangion, A. (2023).
\newblock Likelihood-free neural bayes estimators for censored
  peaks-over-threshold models.
\newblock {\em arXiv preprint arXiv:2306.15642 (2023)}.

\bibitem[Rue et~al., 2009]{rue2009approximate}
Rue, H., Martino, S., and Chopin, N. (2009).
\newblock Approximate {B}ayesian inference for latent {G}aussian models by
  using integrated nested {L}aplace approximations.
\newblock {\em Journal of the Royal Statistical Society: Series B (Statistical
  Methodology)}, 71(2):319--392.

\bibitem[Sainsbury-Dale et~al., 2023]{sainsbury2023neural}
Sainsbury-Dale, M., Richards, J., Zammit-Mangion, A., and Huser, R. (2023).
\newblock Neural bayes estimators for irregular spatial data using graph neural
  networks.
\newblock {\em arXiv preprint arXiv:2310.02600}.

\bibitem[Sainsbury-Dale et~al., 2022]{sainsbury2022fast}
Sainsbury-Dale, M., Zammit-Mangion, A., and Huser, R. (2022).
\newblock Fast optimal estimation with intractable models using
  permutation-invariant neural networks.
\newblock {\em arXiv preprint arXiv:2208.12942}.

\bibitem[Shang et~al., 2015]{shang2015two}
Shang, H., Yan, J., and Zhang, X. (2015).
\newblock A two-step approach to model precipitation extremes in {C}alifornia
  based on max-stable and marginal point processes.
\newblock {\em The Annals of Applied Statistics}, pages 452--473.

\bibitem[Sisson et~al., 2018]{sisson2018handbook}
Sisson, S.~A., Fan, Y., and Beaumont, M. (2018).
\newblock {\em Handbook of approximate Bayesian computation}.
\newblock CRC Press.

\bibitem[Sisson et~al., 2007]{sisson2007sequential}
Sisson, S.~A., Fan, Y., and Tanaka, M.~M. (2007).
\newblock Sequential monte carlo without likelihoods.
\newblock {\em Proceedings of the National Academy of Sciences},
  104(6):1760--1765.

\bibitem[Taylor, 1982]{taylor1982financial}
Taylor, S.~J. (1982).
\newblock Financial returns modelled by the product of two stochastic
  processes-a study of the daily sugar prices 1961-75.
\newblock {\em Time series analysis: theory and practice}, 1:203--226.

\bibitem[Wood, 2010]{wood2010statistical}
Wood, S.~N. (2010).
\newblock Statistical inference for noisy nonlinear ecological dynamic systems.
\newblock {\em Nature}, 466(7310):1102--1104.

\end{thebibliography}

\bibliographystyle{apalike}

\end{document}